# Using First-Order Probability Logic for the Construction of Bayesian Networks

Fahiem Bacchus*
Department of Computer Science
University of Waterloo
Waterloo, Ontario, Canada
N2L–3G1

## Abstract

We present a mechanism for constructing graphical models, specifically Bayesian networks, from a knowledge base of general probabilistic information. The unique feature of our approach is that it uses a powerful first-order probabilistic logic for expressing the general knowledge base. This logic allows for the representation of a wide range of logical and probabilistic information. The model construction procedure we propose uses notions from direct inference to identify pieces of local statistical information from the knowledge base that are most appropriate to the particular event we want to reason about. These pieces are composed to generate a joint probability distribution specified as a Bayesian network. Although there are fundamental difficulties in dealing with fully general knowledge, our procedure is practical for quite rich knowledge bases and it supports the construction of a far wider range of networks than allowed for by current template technology.

## 1 Introduction

The development of graphical representations for probabilistic and decision-theoretic models [Pea88, OS90] has vastly increased the range of applicability of such models in AI. However, it appears that current graphical representations are limited to specialized domains of knowledge—somewhere around the scope of modern expert systems. For a number of reasons, it seems impossible to use such models to represent, say, the general medical knowledge possessed by a typical physician.

A major limitation of current graphical representations is that they are *propositional*. That is, they lack quantifiers, which are essential for representing general knowledge. With quantifiers one can represent an assertion about a whole class of individuals using a single sentence, while in a propositional language this would require a separate sentence for each individual. As a result, important knowledge structuring techniques, like taxonomies, cannot be applied to propositional representations.

*This work was supported by NSERC under its Operating Grants program and by the IRIS network. The author's e-mail address is fbacchus@logos.uwaterloo.ca

However, graphical representations have important advantages of their own. In particular, they support efficient reasoning algorithms. These algorithms are far more efficient than the symbolic reasoning mechanisms typical of more general representations.

This dichotomy of utility has lead to proposals for hybrid uses of general and graphical representations. In particular, Breese et al. [BGW91] have proposed the technique of knowledge based model construction (KBMC): the automatic construction of propositional/graphical models for specific problems from a larger knowledge base expressed in a general representation. Breese et al. provide a number of motivations for this approach that extend the arguments given above.

We refer the reader to [BGW91] for this motivation, and take as our starting point that KBMC is a potentially useful technique, certainly worth examining in more detail. Our contribution, then, is to look more closely at a particular mechanism for performing KBMC. In particular, we develop a mechanism in which a first-order probability logic [Bac90b] is used to represent the general knowledge base, and model construction is performed using ideas arising from the study of *direct inference*. Direct inference involves reasoning from general statistical knowledge to probabilities assigned to particular cases and has been worked on by a number of authors including [BGHK92, Bac90b, Kyb61, Kyb74, Lev80, Lou87, Pol90, Sal71]. Our mechanism brings to light the important role expressive first-order probability logics can play in representing general probabilistic knowledge, and the important relationship between KBMC and direct inference.

In the sequel, we first introduce a probability logic that can be used for the representation of general probabilistic and logical knowledge, and demonstrate that it is capable of representing any Bayesian network [Pea86]—perhaps the most important of current graphical representations. Then we discuss how ideas from direct inference can be used to specify a model construction procedure that can construct graphical models for particular problems. We point out how this idea is related to, but strictly more general than, template models. Throughout our discussion we try to point out various insights about the process of KBMC offered by our approach. Finally, we close with some conclusions and indications for future work.

## 2 Representing General Probabilistic Knowledge

KBMC requires a mechanism for representing general knowledge. This representation should be declarative, for a number of good reasons that are beyond the scope of this paper to discuss. Furthermore, the representation should have a precise semantics, so that we can specify exactly the meaning of the expressions in the knowledge base. Without precise semantics it would be impossible to verify the accuracy of the knowledge base.

Since logical representations meet our desiderata, we propose as a representation mechanism a first-order logic for statistical information, developed by Bacchus [Bac90a]. This logic is basically first-order logic augmented to allow the expression of various assertions about proportions.

Syntactically, we augment an ordinary collection of first-order symbols with symbols useful for expressing numeric assertions, e.g., '1', '+', '$\geq$'. In addition to allowing the generation of ordinary first-order formulas we also allow the generation of numeric assertions involving proportions. For example, $[P(x)]_x = 0.75$, expresses the assertion that 75% of the individuals in the domain have property $P$, while $0.45 \leq [R(x,y)]_{\langle x,y \rangle} \leq 0.55$ expresses the assertion that between 45% and 55% of all pairs of domain individuals stand in relation $R$. In general, if $\alpha$ is an existent formula and $\vec{x}$ is a vector of $n$ variables, the proportion term $[\alpha]_{\vec{x}}$ denotes the proportion of $n$-ary vectors of domain individual that satisfy the formula $\alpha$. Most of the statistical information we wish to express will in fact be statements of conditional probability denoting relative proportions. For example, $[\alpha|\beta]_{\vec{x}}$ will denote the proportion of $n$-ary vectors of domain individuals among those that satisfy $\beta$ which also satisfy $\alpha$. We can then express various statistical assertions by expressing various constraints on the values that these proportion terms can take. For example, by asserting that $[Q(x)|P(x)]_x = 0.5$ we are asserting that the domain we are considering is such that 1/2 of the $P$'s are $Q$'s.

We will not give a formal specification of the semantics of our language here (see [Bac90b] for all such details). The specification simply formalizes the following notion: a formula with free variables might become true or false depending on how the variables are instantiated. For example, $bird(x)$ might be true when $x = Tweety$ but false when $x = Clyde$. A proportion term, then, simply evaluates to the proportion of possible instantiations that make the formula true.

This language can express an wide variety of statistical assertions ([Bac90b] gives an extensive collection of examples). It can also express whatever can be expressed in first-order logic, so essential structuring mechanisms like taxonomies can be applied.

**Example 1** Let the domain contain, among other things, a collection of coins, and a collection of coin-tossing *events*.[1] In addition to some obvious symbols, let our language include the predicate $CoinToss(e)$ which is true of an individual $e$ iff $e$ is an coin-tossing event; $Coin(x)$ which is true of $x$ iff $x$ is a coin; and $Object(e, x)$ which holds of the individuals $e$ and $x$ iff $e$ is an event and $x$ is the object of that event: the object of a coin-tossing event is the particular coin that is tossed. Now we can express the following:

1. $\forall e, x. CoinToss(e) \wedge Object(e, x) \rightarrow Coin(x)$. That is, the object of any coin toss is always a coin.

2. $\forall x. Fair(x) \leftrightarrow [Heads(e)|CoinToss(e) \wedge Object(e,x)]_e \in (.49, .51)$. We agree to call any coin $x$ fair iff approximately 50% of the events in which it is tossed result in heads. This example demonstrates the useful interplay between universal quantification and the proportion terms.

3. $\big[[Heads(e)|CoinToss(e) \wedge Object(e,x)]_e \in (0.49, 0.51) \big| Coin(x)\big]_x = 0.95$. This formula says that 95% of all coins are such that approximately 50% of the events in which they are tossed result in heads. That is, 95% of the coins in the domain are approximately fair. This example demonstrates the useful ability to nest proportion statements.

## 3 Representing Bayesian Networks

Using the logic described in the previous section we can represent a large knowledge base of general logical and statistical information by a collection of sentences. It is not difficult to see that any discrete valued Bayesian network can easily be represented in the logic.[2] Here we will give a particular scheme for representing an arbitrary network, although there are many other schemes possible.

---

[1]The explicit inclusion of *events* in the domain of individuals is similar to the inclusion of other abstract objects like time points or situations (as in the situation calculus [MH69]). There may be philosophical objections, but technical difficulties can be avoided if we restrict ourselves to a finite collection of distinct events.

Any Bayesian network is completely specified by two pieces of information: (1) a product decomposition of the joint distribution which specifies the topological structure of the network, and (2) matrices of conditional probability values which parameterize the nodes in the network [Pea88]. Consider an arbitrary network $B$. Let the nodes in $B$ be the set $\{X_1, \ldots, X_n\}$. Each node $X_i$ has some set of parents $\{X_{f(i,1)}, \ldots, X_{f(i,qi)}\}$, where $f(i,j)$ gives the index of node $X_i$'s $j$-th parent, and $qi$ is the number of parents of $X_i$. Furthermore each node $X_i$ can take one of some discrete set of values $\{v_1, \ldots, v_{ki}\}$, where $ki$ is the number of different values for $X_i$.

The topological structure of $B$ is completely specified by the equation

$$Pr(X_1, \ldots, X_n) = Pr(X_1|X_{f(1,1)}, \ldots, X_{f(1,q1)}) \times \cdots \times Pr(X_n|X_{f(n,1)}, \ldots, X_{f(n,qn)}).$$

That is, the topological structure of $B$ is equivalent to an assertion about how the joint distribution over the nodes $X_1$–$X_n$ can be decomposed into a product of lower-order conditionals. Actually, this equation is shorthand. Its full specification is that this product decomposition holds for every collection of values the nodes $X_1$–$X_n$ can take on.

We can translate this equation into a sentence of our logic by creating a function symbol for every node $X_i$; for convenience we use the same symbol $X_i$. Now the above structure equation can be rewritten as the sentence

$$\forall z_1, \ldots, z_n. [X_1(e) = z_1 \wedge \cdots \wedge X_n(e) = z_n]_e =$$
$$\left[X_1(e) = z_1 \middle| \begin{array}{c} X_{f(1,1)} = z_{f(1,1)} \wedge \cdots \\ \wedge X_{f(1,q1)} = z_{f(1,q1)} \end{array}\right]_e \times$$
$$\vdots$$
$$\times \left[X_n(e) = z_n \middle| \begin{array}{c} X_{f(n,1)} = z_{f(n,1)} \wedge \cdots \\ \wedge X_{f(n,qn)} = z_{f(n,qn)} \end{array}\right]_e.$$

Here we have treated the multi-valued nodes as function symbols $X_1$–$X_n$ in our language. Our translated sentence asserts that for every particular set of values the $X_1$–$X_n$ can take on, the proportion of events $e$ that achieve that set of values can be computed from the lower-order relative proportions. The universal quantification ensures that this product decomposition holds of every collection of values.

Having completely specified the topological structure of $B$, we can equally easily specify the conditional probability parameters in our language. For each node $X_i$, $B$ provides the probability of $X_i$ taking on any of its allowed values under every possible instantiation of its parents $X_{f(i,1)}, \ldots, X_{f(i,qi)}$. This matrix of conditional probabilities consists of a collection of individual equations each of the form

$$Pr(X_i = t_i | X_{f(i,1)} = t_{f(i,1)}, \cdots, X_{f(i,qi)} = t_{f(i,qi)}) = p,$$

where $t_j$ is some value for variable $X_j$, and $p$ is some numeric probability value.

To translate these equations into sentences of our logic we create new constant symbols for every possible value $t_i$ of every node $X_i$; for convenience we use the same symbol $t_i$. Now the above equation can be rewritten as the sentence

$$\left[X_i(e) = t_i \middle| \begin{array}{c} X_{f(i,1)}(e) = t_{f(i,1)} \wedge \cdots \\ \wedge X_{f(i,qi)}(e) = t_{f(i,qi)} \end{array}\right]_e = p.$$

Here we have simply rewritten the conditional probability equations as equations involving the proportion of events in which $X_i$ takes on value $t_i$.

The above procedure can be applied to any network. Thus we make the following observation. *Any discrete valued Bayesian network can be represented as a collection of sentences in the knowledge base.*

What is important to point out about this translation is that the translated assertions represent *template* networks. As pointed out in [BGW91] most probabilistic networks in use in consultation systems are actually template models. That is, the nodes represent generalized events which get instantiated to the particular event under consideration. For example, a node representing "Disease D" will be instantiated to "Patient John R. Smith has disease D," a node representing "Blood test shows low white cell count" will be instantiated to "Blood test *T0906* for patient John R. Smith shows low white cell count," etc. In our representation the template nature of the networks is made explicit: our formulas refer to proportions over classes of similar events not particular events. As we will see this is not a limitation in representational power, rather it is simply a more accurate representation which allows for greater modularity. Propositional networks referring to particular events are to be generated from the knowledge base via model construction techniques.

## 4  Simple Model Construction

To introduce the basic ideas that underlie our model construction technique consider a knowledge base that consists simply of a collection of template Bayesian networks, each one applicable to different types of events.

To specify that each different decomposition, and collection of conditional probability parameters, is applicable to a different class of events we only need add the event type as an extra conditioning formula. For example, say that we have two networks both suitable for diagnosing abdominal pain. However, one of the networks is designed for women in late-term pregnancy, while the other is suitable for other patients with abdominal pain. Our general knowledge base might contain the two formulas (Equations 1 and 2) shown in Figure 1.

---

[2]It is also possible, with a few technical caveats, to represent networks with continuous valued nodes. But here we restrict our attention to discrete valued nodes.

(1)
$$\forall z_1, z_2, z_3.[X_1(e) = z_1 \wedge X_2(e) = z_2 \wedge X_3(e) = z_3 | AbdominalPain(e) \wedge \neg Pregnancy(e)]_e$$
$$= [X_1(e) = z_1 | AbdominalPain(e) \wedge \neg Pregnancy(e)]_e$$
$$\times [X_2(e) = z_2 | X_1(e) = z_1 \wedge AbdominalPain(e) \wedge \neg Pregnancy(e)]_e$$
$$\times [X_3(e) = z_3 | X_1(e) = z_1 \wedge X_2(e) = z_2 \wedge AbdominalPain(e) \wedge \neg Pregnancy(e)]_e,$$

(2)
$$\forall z_1, z_2, z_3.[Y_1(e) = z_1 \wedge Y_2(e) = z_2 \wedge Y_3(e) = z_3 | AbdominalPain(e) \wedge Pregnancy(e)]_e$$
$$= [Y_1(e) = z_1 | AbdominalPain(e) \wedge Pregnancy(e)]_e$$
$$\times [Y_2(e) = z_2 | Y_1(e) = z_1 \wedge AbdominalPain(e) \wedge Pregnancy(e)]_e$$
$$\times [Y_3(e) = z_3 | Y_1(e) = z_1 \wedge AbdominalPain(e) \wedge Pregnancy(e)]_e.$$

Figure 1: Alternate Structures for Abdominal Pain

In this example the events involving abdominal pain and pregnancy have a different network models (i.e., structural decompositions) with entirely different variables than the events where there is no pregnancy. In a similar manner we can represent a whole collection of disjoint types of events, where each event type is modeled by a different probabilistic structure.

In this case the model construction technique in this case would simply locate the appropriate template model using information about the particular event being reasoned about. For example, if the event is $E001$ and we know $AbdominalPain(E001) \wedge Pregnancy(E001)$, i.e., the event being reasoned about involves adominal pain in a pregnant patient, we would construct a network model for reasoning about $E001$ using the second template model. This network would have the structure

$$Pr(Y_1, Y_2, Y_3) = Pr(Y_1) \times Pr(Y_2|Y_1) \times Pr(Y_3|Y_1),$$

and would be parameterized by the values specified in the knowledge base for the $Y_i$ variables. Since the constructed network is now specific to event $E001$ we can drop the extra condition $AbdominalPain(e) \wedge Pregancy(e)$ as we know that $E001$ satisfies these conditions. Now we have an event specific network that can be used to reason about the probable values of the variables $Y_i$ in the particular event.

We can see that the model constructor is simply "instantiating" the general template model with the particular event $E001$. By using the same structure and probability parameters as the class of abdominal pain-pregnancy events we are assigning probabilities to the particular event $E001$ that are identical to the statistics we have about that general class of events. This is an example of direct inference, where we use statistics over a class of similar events to assign probabilities to a particular event. For example, when we assign a probability of 1/2 to the event of heads on a *particular* coin toss based on statistics from a *series* of coin tosses we are performing direct inference. This kind of inference is pervasive in reasoning under uncertainty.[3]

Simple model construction of this kind is not that interesting however. We could easily accomplish the same thing with a control structure that chooses from some collection of networks. The main difference is that here we have an explicit, declarative, representation of which network is applicable to what type of event. Furthermore, it also serves to illustrate the basic idea behind our approach to KBMC.

## 5 More General Model Construction

In general we will not have explicit template models in our knowledge base for all of the events we wish to reason about. Indeed, this is exactly the point of the KBMC approach: we want to deal with situations beyond the ability of template models.

Our knowledge base will more likely contain information about conditional probabilities isolated to neighborhoods of related variables. For example, instead of having an explicit product decomposition for all of the relevant variables, as in the above examples, the knowledge base might simply contain the individual product terms, i.e., the neighborhood information, in isolation. It will be up to the model construction procedure to link these individual terms into a joint distribution. Consider Pearl's classic Holmes's burglary example. It is unlikely that Holmes has in his knowledge base an explicitly represented decomposition of the form shown in Equation 3 (Figure 2). Such a decomposition is simply far too specific. Rather Holmes would more typically have information like that shown in Equation 4 (Figure 2). In this case Holmes has the knowledge (a) in 75% of the events in which a house with an alarm is burglarized, the alarm will sound; (b) in 45% of the events in which an alarm sounds near where a person lives that person will report the alarm; (c) the specific knowledge that Watson lives near Holmes's house and that Holmes's house has an alarm. The advantage of knowledge in this more general form is that it can be used to reason about many other types of events. For example, the statistical knowledge (a) can be used to reason about any alarm in any house, e.g., if Holmes learns that his parents house alarm has been tripped; similarly (b) can be

$$
\begin{aligned}
(3) \quad & [Burglary(e, MyHouse) \wedge AlarmSound(e, MyHouse) \wedge ReportsAlarm(e, Watson, MyHouse)]_e \\
& = [AlarmSound(e, MyHouse)|Burglary(e, MyHouse)]_e \\
& \quad \times [ReportsAlarm(e, Watson, MyHouse)|AlarmSound(e, MyHouse)]_e.
\end{aligned}
$$

$$
\begin{aligned}
(4) \quad (a) \quad & [AlarmSound(e, x)|Burglary(e, x) \wedge HouseWithAlarm(x)]_{\langle e, x \rangle} = .75 \\
(b) \quad & [ReportsAlarm(e, y, x)| \\
& \quad AlarmSound(e, x) \wedge HouseWithAlarm(x) \wedge LivesNear(x, y)]_{\langle e, x, y \rangle} = 0.45 \\
(c) \quad & LivesNear(MyHouse, Watson) \wedge HouseWithAlarm(MyHouse)
\end{aligned}
$$

---

[3] See Kyburg [Kyb83a] for further arguments pointing out the prevalence of "direct inference" in probabilistic reasoning.

Figure 2: An overly Specific Decomposition vs. General Information

used for reasoning about reports from any neighbor, e.g., if Mrs. Gibbons reported the alarm instead of Dr. Watson.

Holmes will also have other pieces of statistical information, e.g., statistics about the event that a house has been burglarized given that a police car is parked outside, and other pieces of information specific to the particular event being reasoned about. The task, then, of a model construction procedure is to use the information specific to the particular event being reasoned about to decide which local pieces of statistical information are relevant and how they should be linked into a Bayesian network representation. Once a network has been constructed it can be used to quickly perform a range of complex reasoning about the particular event.

There are three issues that arise when constructing a Bayesian network model of the particular event we are reasoning about. First, the model construction procedure must have some information about the variables (properties of the event in question) that we wish to include in the constructed network. Second, we must use information about the particular event to locate appropriate pieces of local statistical information in the knowledge base. And third, we must combine these local pieces of information into a network.

### 5.1 The Set of Variables

Some information must be supplied about what collection of variables we want to model in the constructed network. In the simplest case we will just supply a query about the particular event under consideration along with some additional information about that event. For example, we might be reasoning about event $E002$ and the query might be expressed as $Burglary(E002)$?; i.e., did a burglary occur as part of this event? We might also have the information $ReportsAlarm(E002, Watson, MyHouse)$, i.e., Dr. Watson reported an alarm at Holmes's house during this event. If the knowledge base is similar to that given above, the procedure could determine that it can chain probabilistic influence from a report by Watson to belief in the alarm sounding, and then from there to a belief in a burglary, i.e., to an inference about the query. Given that this is the only chain of influence it can find in the knowledge base linking alarm reports and burglaries, the constructed network will only contain a burglary node, an alarm sound node, and an alarm report node. That is, in a strictly query driven KBMC procedure the constructed model will only contain variables relevant to the particular query.

Alternately, we could supply the procedure with more information. For example, we could specify a set of variables that we wish to include in the constructed model. For example, we could specify that we are also interested in reasoning about earthquakes and radio broadcasts. If the knowledge base has local statistics about the frequency of alarms sounding given earthquakes, and radio reports given earthquakes, a larger Bayesian network could be constructed that includes nodes for these variables. The links between these variables would be determined by the local statistics contained in the knowledge base. For example, if we know the frequency of alarm triggers given earthquake events, we would place a link from the earthquake node to the alarm node in the constructed network.

As in the simple query driven case, however, the procedure would still be able to add additional intermediate variables that link the variables in the set of interest. These intermediate variables would be found by looking through the knowledge base for chains of influences between the specified variables. For example, if we inform the procedure to build a model of some set of diseases $\{D_1, \ldots, D_n\}$ and some set of symptoms $\{S_1, \ldots, S_m\}$, it can search for *chains* of local conditional probabilities linking members of these two sets. Hence, the constructed network will generally contain additional intermediary nodes describing the causal processes known to link the diseases with the various symptoms, just as the alarm sound information linked burglaries and alarm reports in the query driven case.

It seems likely that we would want to amortize the effort of constructing the Bayesian network over a whole range of queries. Hence, we will probably want to supply the model constructor with more information than just a single query.

### 5.2 Locating the Appropriate Local Statistics

Information about the particular event will help determine which collection of local statistics are appropriate. The issue of choosing appropriate statistics is at the heart of the difficulties in direct inference. Ol-approaches to direct inference revolved around trying to find appropriate reference classes from which statistics can be drawn [Kyb83b]. More recent work has taken an approach based on the principle of indifference that dispenses with the notion of a reference class altogether [BGHK92]. In general, however, determining the probabilities to assign to a particular event given a collection of statistical information about classes of similar events is a very difficult problem. For a practical enterprise like KBMC, however, we can use the work on direct inference to derive general guidelines as to what statistics to consider. For example, all approaches to direct inference validate the subset or specificity preference: one should choose the most specific statistics applicable to the event in question. Similarly, if we have statistical information about a specific individual involved in the event we should use that.

Information about the particular event can alter both the parameterization and the structure of the constructed Bayesian network. This flexibility is not possible with simple template models. Consider the following example.

**Example 2** Say that the local information shown in Figure 3 was contained in the knowledge base. And say that our information about the particular event was *ReportsAlarm(E002, Watson, MyHouse)*. If it is decided that *AlarmSound* should be placed in the constructed network, either because it is a variable of interest or because it is in a chain of influences to a variable of interest, then the procedure would have to choose how to parameterize the link from the *MyHouse* alarm sound node and the Watson alarm report node.

The only statistic we have about the chance of an alarm report given an alarm concerns the class of people who live near the house whose alarm sounded. In this case we know Dr. Watson is a member of this class, i.e., *LivesNear(MyHouse, Watson)*, so item 1 gives the most specific known probability of a report given an alarm. However, we do have a more specific statistic for Dr. Watson, item 3, in the case of a report when there is no alarm, indicating that Watson is a bit of a practical joker. Hence, this more specific value would be used for the probability of a report given no alarm. On the other hand if the event in question involved a report by Mrs. Gibbons, we would be forced to use the more general statistics 1 and 2 to parameterize the alarm-report/alarm-sound link as we have no specific statistics for Mrs. Gibbon's alarm reports.

**Example 3** Let the knowledge base be as in Figure 3, except augmented by the additional statistical information shown in Figure 4. That is, in this case Holmes has a special alarm installed by a security company *AlarmMonitorCompany* with a direct line to their office, and from the company's literature about the accuracy of their alarm systems Holmes has come to accept the above statistical assertion about the reliability of their alarm reports. Now if the event was *ReportsAlarm(E003, AlarmMonitorCompany, MyHouse)* there would be no need for the model construction procedure to include an intermediary node of alarm sound, nor would the direction of the links be required to go from burglaries towards alarm reports. Instead it could use this statistic, as the particular event *E003* is a member of this class of events, to link the alarm report node directly to the burglary node, and a quite different network structure would result.

## 5.3   Linking the Local Pieces

Once appropriate local statistics are obtained from the database we have enough information to link various nodes in the network. That is, each local statistic will serve to parameterize a link between two nodes in the network. A difficulty that arises here is justifying this composition.

All we really know about the probability distribution describing the interaction between the variables are the local conditional probabilities. There will in general be many different joint probability distributions that are consistent with these local conditional probabilities. In linking up the nodes in a manner determined solely by the local information we are constructing a particular joint distribution, one in which the local conditional probabilities determine a product decomposition. An important question is: to what extent is such a procedure justified? Lewis [LI59] proved some results which show that by taking the product of local conditional probabilities one obtains a best estimator in the sense of Kullback-Leibler cross-entropy [KL51]. But his results do not cover all of the cases that might occur. Another justification comes from recent work that applies the principle of indifference to reasoning about change [BGHK93]. For an enterprise like KBMC, however, we will again want to use general principles derived from such work. One general principle arising from [BGHK93], and earlier work by Hunter [Hun89], is that when the variables are causally related, as compared to being simply correlated, using the product of the local conditional probabilities can be justified by principles of indifference.

A related difficulty occurs when we have some but not all of the information required to specify the parameterization of the network. For example, we might have statistics about a number of distinct causes for an effect, but we might not have statistics about their joint effect. Pearl [Pea88] has suggested the use of "prototypical structures" like noisy OR gates. There is an underlying probabilistic model from which noisy OR gates arise, and when it is reasonable to assume that this model holds in a domain, prototypical structures of this form could be used. Alternately, the indifference considerations of [BGHK93, Hun89] can also be used in certain cases to complete the joint distribution over the different causes.

1. $[ReportsAlarm(e,y,x)|$
   $\quad AlarmSound(e,x) \land HouseWithAlarm(x) \land LivesNear(x,y)]_{\langle e,x,y \rangle} = 0.45$
2. $[ReportsAlarm(e,y,x)|$
   $\quad \neg AlarmSound(e,x) \land HouseWithAlarm(x) \land LivesNear(x,y)]_{\langle e,x,y \rangle} = 0.05$
3. $[ReportsAlarm(e,Watson,x)|$
   $\quad \neg AlarmSound(e,x) \land HouseWithAlarm(x) \land LivesNear(x,Watson)]_{\langle e,x \rangle} = 0.15$
4. $HouseWithAlarm(MyHouse) \land LivesNear(MyHouse,Watson)$
5. $LivesNear(MyHouse,Gibbons)$

Figure 3: Knowledge Base for Example 2

6. $[Burglary(e,MyHouse)|ReportsAlarm(e,AlarmMonitorCompany,MyHouse)]_e = 0.90$
7. $[Burglary(e,MyHouse)|\neg ReportsAlarm(e,AlarmMonitorCompany,MyHouse)]_e = 0.05$

Figure 4: Additional Knowledge for Example 3

## 6 Conclusions and Future Work

We have outlined a mechanism for KBMC of Bayesian networks from a knowledge base expressed in a first-order probabilistic logic. Although we have only been able to present a sketch of how the mechanism works we have discussed the main ideas behind the proposal: (1) identify the variables of interest either through a query driven process or through information provided by the user; (2) locate local statistics, relevant to the particular event being reasoned about, by using principles from work on direct inference, like specificity, to prefer certain local statistics over others; (3) construct chains of probabilistic influence from these local statistics; (4) construct an event specific network by using the chains of probabilistic influence to specify the arcs in the network, and by using the local statistics to parameterize the nodes, perhaps filling in missing parameters by using prototypical structures or principles of indifference. The resulting network can then be used to reason probabilistically about the particular event.

The mechanism can be actualized fairly easily in straightforward cases. In such cases the chains of influence are easy to locate: the individual links are explicitly expressed in the knowledge base. If the statistics in the knowledge base are of a form such that selecting the most appropriate statistics reduces to simple specificity considerations and if we have sufficient statistical information, we can easily parameterize the resulting structure. Such a mechanism, although limited in some ways, already offers a considerable increase in flexibility over current template models.

One issue we have not addressed here is a mechanism for representing temporal information, but as shown by Bacchus et al. [BTH91] first-order logic is sufficient for representing a range of temporal ontologies. Hence, once an appropriate temporal ontology is decided upon, it is possible that the representation could be extended to allow for temporal information. If the temporal structure is discrete we could also allow the formation of proportion statements over time points, thus allowing the expression of various assertions about discrete stochastic processes. A related issue that can be addressed is the representation of utilities. Extending our representation to utilities and temporal information, and the KBMC procedure we proposed to generate, e.g., influence diagrams, is an interesting area for future research. Current work on this model is focused on filling in the details of the mechanism we have sketched, and on building a prototype system.

In conclusion, we feel that our proposal is a workable one, that, with sufficient resources, can be turned into a prototype implementation. Work on this is continuting. Such an implementation holds the promise of a useful KBMC procedure that would be far more general than current template models. There are, of course, limitations to the approach, limitations that stem mainly from problems that arise during direct inference. Given a very general knowledge base of statistical information it will not always be possible to choose the "most appropriate" statistical information for an event. For example, we might have conflicting statistical information that cannot be resolved by specificity. Nevertheless, we can still obtain useful results in less general but, we hope, still practical, contexts.

## References

[Bac90a]  Fahiem Bacchus. Lp, a logic for representing and reasoning with statistical knowledge. *Computational Intelligence* 6(4):209–231, 1990.

[Bac90b]  Fahiem Bacchus. *Representing and Reasoning With Probabilistic Knowledge* MIT-Press, Cambridge, Massachusetts 1990.

[BGHK92]  F. Bacchus, A. J. Grove, J. Y. Halpern, and D. Koller. From statistics to belief. In

*Proceedings of the AAAI National Conference*, pages 602–608, 1992.

[BGHK93] F. Bacchus, A. J. Grove, J. Y. Halpern, and D. Koller. Forming beliefs about a changing world. In preparation, 1993.

[BGW91] John S. Breese, Robert Goldman, and Michael P. Wellman. Knowledge-based construction of probabilistic and decision models: An overview. Unpublished manuscript, presented as introduction to AAAI-91 Workshop on Knowledge Based Model Construction. Available from Michael P. Wellman, USAF Wright Laboratory, Wright Patterson Air Force Base, OH., 1991.

[BTH91] Fahiem Bacchus, Josh Tenenberg, and Koomen Hans. A non-reified temporal logic. *Artificial Intelligence*, 52:87–108, 1991.

[Hun89] D. Hunter. Causality and maximum entropy updating. *International Journal of Approximate Reasoning*, 3(1):379–406, 1989.

[KL51] S. Kullback and R. A. Leibler. Information and sufficiency. *Annals of Mathematical Statistics*, 22:79–86, 1951.

[Kyb61] Henry E. Kyburg, Jr. *Probability and the Logic of Rational Belief*. Wesleyan University Press, Middletown, Connecticut, 1961.

[Kyb74] Henry E. Kyburg, Jr. *The Logical Foundations of Statistical Inference*. D. Reidel, Dordrecht, Netherlands, 1974.

[Kyb83a] Henry E. Kyburg, Jr. *Epistemology and Inference*. University of Minnesota Press, 1983.

[Kyb83b] Henry E. Kyburg, Jr. The reference class. *Philosophy of Science*, 50(3):374–397, September 1983.

[Lev80] Isaac Levi. *The Enterprise of Knowledge*. MIT-Press, Cambridge, Massachusetts, Cambridge, Massachusetts, 1980.

[LI59] P. M. Lewis II. Approximation probability distributions to reduce storage requirements. *Information and Control*, 2:214–225, 1959.

[Lou87] Ronald P. Loui. *Theory and Computation of Uncertain Inference and Decision*. PhD thesis, The University of Rochester, September 1987.

[MH69] John McCarthy and Patrick J. Hayes. Some philosophical problems from the standpoint of artificial intelligence. In *Machine Intelligence 4*, pages 463–502. Edinburgh University Press, 1969.

[OS90] Robert M. Oliver and John Q. Smith, editors. *Influence Diagrams, Belief Nets and Decision Analysis*. Wiley, 1990.

[Pea86] Judea Pearl. Fusion, propagation, and structuring in belief networks. *Artificial Intelligence*, 29:241–288, 1986.

[Pea88] Judea Pearl. *Probabilistic Reasoning in Intelligent Systems*. Morgan Kaufmann, San Mateo, California, 1988.

[Pol90] John L. Pollock. *Nomic Probabilities and the Foundations of Induction*. Oxford University Press, Oxford, 1990.

[Sal71] Wesley Salmon. *Statistical Explanation and Statistical Relevance*. University of Pittsburgh Press, Pittsburgh, 1971.